\documentclass{article}
\usepackage{iclr2022_workshop,times}

\usepackage{amsmath,amsfonts,bm}

\def\eqref#1{equation~\ref{#1}}

\def\1{\bm{1}}

\DeclareMathAlphabet{\mathsfit}{\encodingdefault}{\sfdefault}{m}{sl}
\SetMathAlphabet{\mathsfit}{bold}{\encodingdefault}{\sfdefault}{bx}{n}

\DeclareMathOperator*{\argmin}{arg\,min}

\usepackage{wrapfig}
\usepackage{multicol}
\usepackage{hyperref}
\usepackage{url}
\usepackage{tikz}
\usepackage{algorithm}
\usepackage{algpseudocode}

\usepackage{xcolor}

\title{Fiber Bundle Morphisms as a Framework for Modeling Many-to-Many Maps}

\author{Elizabeth Coda, Nico Courts, Colby Wight, Loc Truong, WoongJo Choi, \\\textbf{Charles Godfrey, Tegan Emerson\thanks{ Dr. Emerson holds joint appointments in the Department of Mathematics at Colorado State University and the Department of Mathematical Sciences at the University of Texas, El Paso.}, Keerti Kappagantula, Henry Kvinge\thanks{Dr. Kvinge holds a joint appointment in the Department of Mathematics at the University of Washington.}} \\
Pacific Northwest National Lab\\
\texttt{\{first.last\}@pnnl.gov} \\
\AND 
}

 \iclrfinalcopy 
\begin{document}

\maketitle

\begin{abstract}
While it is not generally reflected in the `nice' datasets used for benchmarking machine learning algorithms, the real-world is full of processes that would be best described as many-to-many. That is, a single input can potentially yield many different outputs (whether due to noise, imperfect measurement, or intrinsic stochasticity in the process) and many different inputs can yield the same output (that is, the map is not injective). For example, imagine a sentiment analysis task where, due to linguistic ambiguity, a single statement can have a range of different sentiment interpretations while at the same time many distinct statements can represent the same sentiment. When modeling such a multivalued function $f: X \rightarrow Y$, it is frequently useful to be able to model the distribution on $f(x)$ for specific input $x$ as well as the distribution on fiber $f^{-1}(y)$ for specific output $y$. Such an analysis helps the user (i) better understand the variance intrinsic to the process they are studying and (ii) understand the range of specific input $x$ that can be used to achieve output $y$. Following existing work which used a fiber bundle framework to better model many-to-one processes, we describe how morphisms of fiber bundles provide a template for building models which naturally capture the structure of many-to-many processes.
\end{abstract}

\section{Introduction}

Variation is ubiquitous in the real-world. This is especially true for physical processes where a single input to the system can produce many different possible outputs. For example, in nearly all scientific fields it is common for an experiment to yield (slightly) different results each time it is repeated, even when all controllable parameters are held fixed. On the other hand, variation in systems can also flow in the other direction. It is common to have many inputs to the system that yield a single output. For example, there are infinitely many images of cats that all map to the label `cat'. From the perspective of machine learning (ML) we can think of processes that possess both of these properties as being many-to-many. Each input to the system may yield many different outputs and many different inputs can also yield a single output. A non-injective multivalued function underlies such processes.

In this paper we develop a deep learning architecture capable of modeling the distributions found on both images and fibers of a many-to-many map. More precisely, let $f:X \rightarrow Y$ be a many-to-many map, so that $f(x)$ and $f^{-1}(y)$ are both sets for any $x \in X$ and $y \in Y$. A probability distribution on $X$ induces probability distributions on both $f(x)$ and $f^{-1}(y)$. Our goal is to build a model that simultaneously learns all of these distributions over the course of training. Such information may not be necessary for simple inference tasks, but modeling the distributions on both images and fibers gives information that can be critical for decision making. For example, understanding the distribution $f^{-1}(y)$ can help a user identify all possible inputs that can be applied to achieve the specific desired output $y$. On the other hand, being able to understand $f(x)$ for varying $x$ can point to those $x$ that may yield less consistent values in $Y$.  

The present work was inspired by a problem in advanced manufacturing where one wants to understand all the different manufacturing processing parameter settings $X$ that will lead to a single desirable material property $y \in Y$. If $f$ represents the manufacturing process this amounts to understanding $f^{-1}(y)$. At the same time, certain processing parameters in $X$ can yield less inconsistent results (more variation in $f(x)$), which is often useful information when each processing run is time-, labor-, and resource-intensive.

To achieve both of these objectives within a single model architecture we build off the work of \citet{courts2021bundle} which proposed a fiber bundle framework to model many-to-one maps. We find that while fiber bundles offer a useful template for many-to-one maps, bundle morphisms better capture the structure required for an architecture meant to capture many-to-many maps. In the model that we propose, which we call a {\emph{Bundle Morphism Network (BMNet)}}, we offload the variability intrinsic to $f$ by lifting it to a new `bundle morphism’ $F: E_X \rightarrow E_Y$. We incorporate local trivializations into $F$ so that it is easy to move from $X$ to $E_X$ and $Y$ to $E_Y$. Finally, the variation that emerges when we consider either $f(x)$ or $f^{-1}(y)$ is encoded in simple distributions (e.g., 1-dimensional Gaussians) on the fibers of $E_X$ and $E_Y$ respectively. The local trivializations, which we build into our model using invertible neural networks, are trained to transform these distributions to the distribution on $f(x)$ and $f^{-1}(y)$.

We test BMNet on several synthetic datasets that represent many-to-many maps with a fiber bundle like flavor. We benchmark BMNet against other many-to-many models as well as their relatives such as conditional GANs (cGANs) \citep{mirza2014conditional} and conditional normalizing flow (cNF) models \citep{winkler2019}. We show that BMNet generally outperforms these models on most of our synthetic datasets suggesting that the next step is to benchmark BMNet against a wider variety of more complicated real-world datasets.

\section{Fiber Bundles and Bundle Morphisms}

Early in the history of geometry it was recognized that one way to analyze a space is to decompose it into simpler constituent parts. The most familiar of such decompositions is the product, wherein space $X$ is identified as the Cartesian product of two simpler spaces $Y$ and $Z$, $X \cong Y \times Z$. The torus $T$ is a simple example of a space that is (topologically) the product of two spheres, $T \cong S^1 \times S^1$. Some spaces, however, may appear to be a product space locally and yet fail to be a product space globally due to a `twist' in the product structure. One of the simplest examples of this is the M\"obius band. The notion of a fiber bundle, first introduced in \citet{Seifert} and \citet{Whitney}, is meant to capture the essence of this phenomenon. We include a review of the definition of a fiber bundle in Section \ref{sect-fiber-bundles} of the Appendix.

\begin{wrapfigure}{r}{0.5\linewidth}
\vspace{-.3cm}
\label{eqn-comm-diagram}
\centering
    \begin{tikzpicture}
    \node at (0,0) {$E_1$};
    \node at (2.5,0) {$E_2$};
    \node at (0,-1.2) {$B_1$};
    \node at (2.5,-1.2) {$B_2$};
    
    \draw[->,black] (.3,0)--(2.2,0);
    \draw[<-,black] (2.2,-1.2)--(.3,-1.2);
    \draw[->,black] (0,-.3)--(0,-.9);
    \draw[->,black] (2.5,-.3)--(2.5,-.9);
    
    \node at (1.3,.28) {\scriptsize{$F$}};
    \node at (1.3,-.86) {\scriptsize{$f$}};
    \node at (-.2,-.6) {$\pi_1$};
    \node at (2.8,-.6) {$\pi_2$};
    \end{tikzpicture}
    \caption{Bundle morphism diagram.}
\end{wrapfigure}

As with many structures within topology and geometry, it is useful to identify the maps between fiber bundles that preserve their structure. Let $\mathbf{E_1} = (E_1,B_1,Z_1,\pi_1)$ and $\mathbf{E_2} = (E_2,B_2,Z_2,\pi_2)$ be two fiber bundles. A {\emph{bundle morphism}} $F$ between total spaces $\mathbf{E_1}$ and $\mathbf{E_2}$ is a continuous map $F: E_1 \rightarrow E_2$ and continuous map on base spaces $f: B_1 \rightarrow B_2$ such the following diagram commutes.

While it is not intended to follow this definition ``on the nose'', the concept of a bundle morphism informs the structure of the Bundle Morphism Network introduced in Section \ref{sect-algo-description}.


\section{Bundle Morphism Network}
\label{sect-algo-description}

\begin{figure}[h]
    \centering
    \includegraphics[ scale=0.26]{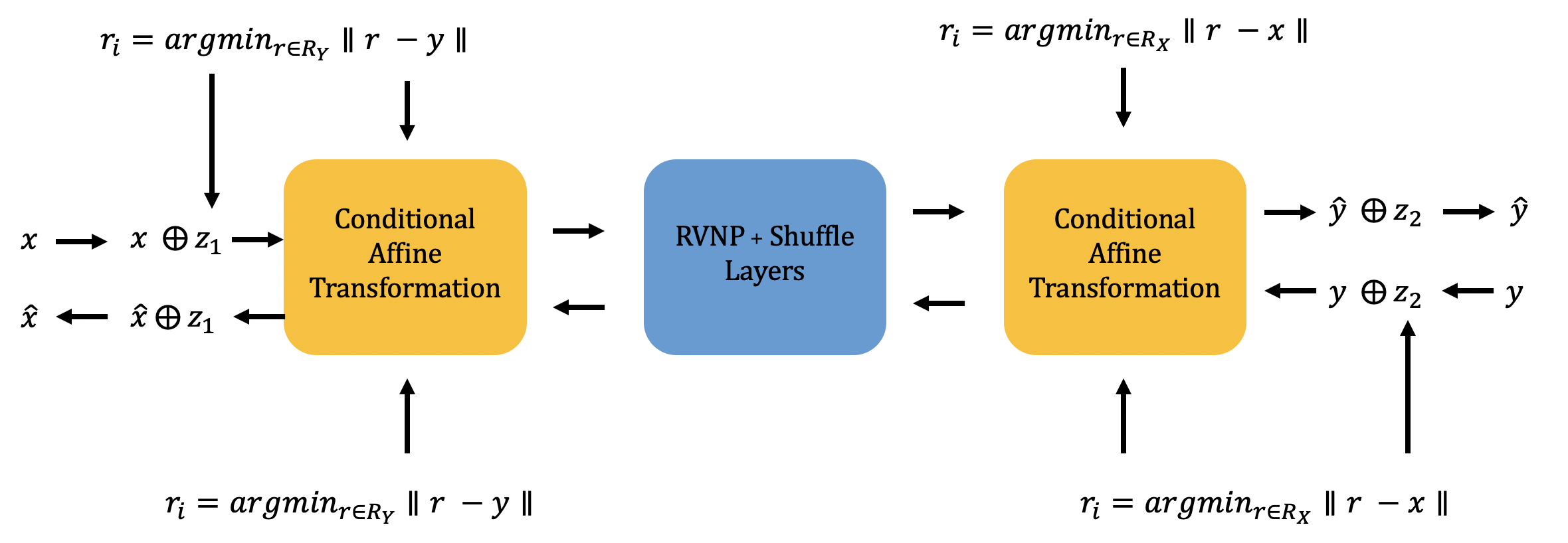}
   \caption{A diagram outlining the architecture of Bundle Morphism Network.}
   \label{fig:architectyre}
\end{figure}

Building off of BundleNet, {\emph{Bundle Morphism Network (BMNet)}} is a neural network $\Phi: X \times Z_{1} \times \mathcal{R}_X \times \mathcal{R}_Y \rightarrow  Y \times Z_{2}$ where $X$ is the input space for the task, $Y$ is the output space, $Z_1$ and $Z_2$ are called the fibers, and $\mathcal{R}_X = \{r_1^X, \dots, r_{n_1}^X\}$ and $\mathcal{R}_Y = \{r_1^Y, \dots, r_{n_2}^Y\}$ are a collection of conditioning vectors representing neighborhoods of $X$ and $Y$ respectively. For fixed $r_i^X \in \mathcal{R}_X, r_j^Y \in \mathcal{R}_Y$ we write $\Phi_{i,j} := \Phi(-,-,r_i^X,r_j^Y)$. $\Phi_{i,j}$ is invertible (we use the \texttt{FrEIA} package \citep{FrEIA} to build $\Phi$ in our experiments). For clarity we refer to the process of running the model in the direction $X \times Z_{1} \rightarrow  Y \times Z_{2} $ as the {\emph{forward direction}} and running the model in the direction  $Y \times Z_{2} \rightarrow  X \times Z_{1} $ as the {\emph{reverse direction}}. $Z_1$ and $Z_2$ are assumed to follow user chosen distributions $\mathcal{D}_1$ and $\mathcal{D}_2$. In the experiments in this paper we choose to use the uniform distribution on the circle when sampling from $Z_1$ and $Z_2$ (justification for this was given in \citep{courts2021bundle}).

Elements of $\mathcal{R}_X$ and $\mathcal{R}_Y$ consist of representatives of neighborhoods in the support of the training data input distribution (in $X$) and the training data output distribution (in $Y$) respectively. Given a training set $D \subset X \times Y$, we cluster both the input from $X$ and the labels from $Y$ and let $\mathcal{R}_X$ and $\mathcal{R}_Y$ be the cluster centroids. Described in the forward direction, the model consists of a conditional affine transformation conditioned on elements of $\mathcal{R}_{Y}$, several RVNP blocks and coordinate permutation layers \citep{RNVP}, and an additional conditional affine transformation conditioned on elements of $\mathcal{R}_{X}$. These invertible conditional layers correspond to the local trivializations for each ``bundle’’.

During training, a batch consists of training examples $\{(x_{i}, y_{i})\}$ such that all $x$ in the batch belong to the same cluster with center $r_{i}^X \in  R_{X}$ and all $y$ in the batch belong to the same cluster with center $r_{j}^Y \in  R_{Y}$. For each $x$, we sample a $z_{1}$ from distribution $\mathcal{D}_1$ on $Z_1$. To run the model in the forward direction, we condition on $r_{j}^Y$ and $r_{i}^X$ and obtain $(\hat{y}, \hat{z}_{2}) = \Phi(x, z_{1}, r_{i}^X, r_{j}^Y)$. To run the model in the reverse direction, we sample $z_{2}$ and obtain $(\hat{x}, \hat{z}_{1}) = \Phi^{-1}(y, z_{2},r_{i}^X, r_{i}^Y)$. Once we have a collection of such $\hat{y}$, which we denote by $\hat{S}$, and $y$, which we denote by $S$, the loss in the forward direction is  

\begin{equation*}
\mathcal{L}_{\text{forward}}(\hat{S}, S) = \frac{1}{|\hat{S}|} \sum_{\hat{s} \in \hat{S}} \min_{ s \in S}  || \hat{s} - s || ^{2} + \frac{1}{|S|} \sum_{s \in S} \min_{ \hat{s} \in \hat{S}}  || \hat{s} - s || ^{2}.
\end{equation*}

$\mathcal{L}_{\text{reverse}}$ is defined analogously. This is a symmetric version of the mean-squared minimum distance, used to ensure that the learned distribution covers the entirety of the training distribution. Notice that this is different from the loss function in \cite{courts2021bundle}. We found that applying an analogous version of the loss found in that paper was unstable when applied to our problem. The detailed training algorithm is outlined in Algorithm \ref{alg:train}. 

At inference time, we will only have access to one of $x$ or $y$. If we wish to run the model in the forward direction and generate the distribution of $f(x)$, we need a conditioning vector $r_{j}^Y \in  \mathcal{R}_{Y}$, which we cannot assume we have since we do not have a $y$ value paired with $x$. To get around this problem, we find the $k$ nearest neighbors of $x$ from the training set and randomly sample one $(x',y')$. We use the corresponding $y'$ from $(x',y')$ to determine $r_{j}^Y$. The detailed inference algorithm is outlined in Algorithm \ref{alg:inference}.  Running inference in the reverse direction is analogous. 

\begin{algorithm}[t]
\caption{Training Procedure}\label{alg:train}

\textbf{Input:} Training dataset $D = \{ (x_i, y_i) \}$, prior distributions $\mathcal{D}_1$ and $\mathcal{D}_2$  \\
\textbf{Output:} Trained model $\Phi$ with cluster centers $\mathcal{R}_X$ and $\mathcal{R}_Y$ and updated priors $\mathcal{D}_1$ and $\mathcal{D}_2$ 
\begin{algorithmic}[1]
\State Cluster the training data (in $X$) to obtain $n_1$ cluster centers $\mathcal{R}_X$
\State Cluster the training data (in $Y$) to obtain $n_2$ cluster centers $\mathcal{R}_Y$
\For{each epoch}
\For{$r_i^X \in \mathcal{R}_X$ and $r_j^Y \in \mathcal{R}_Y$ }
\State Construct $D_{ij} = \{ (x,y) | r_i^X =  \argmin_{r \in \mathcal{R}_X} |r - x |, r_j^Y =  \argmin_{r \in \mathcal{R}_Y} |r - y |\}$ 
\State $X = \text{proj}_{x}({D_{ij}})$ (set projection to $1$st coordinate)
\State $Y = \text{proj}_{y}({D_{ij}})$ (set projection to $2$nd coordinate)
\State Sample $Z_{1} \sim   \mathcal{D}_1^{i}$ and $Z_{2} \sim   \mathcal{D}_2^{j}$
\State $ ( \hat{Y}, \hat{Z}_{2}) = \Phi(X, Z_{1}, r_{i}^X, r_{j}^Y)$
\State $ ( \hat{X}, \hat{Z}_{1}) = \Phi^{-1}(Y, Z_{2}, r_{i}^X, r_{j}^Y)$
\State $\mathcal{L} = \mathcal{L}_{forward}( \hat{Y}, Y) + \mathcal{L}_{reverse}( \hat{X}, X)$
\State Update the weights of $\Phi$ 
\State Update the parameters of the prior distributions $\mathcal{D}_1^{i}$ and $\mathcal{D}_2^{j}$ 

\EndFor
\EndFor

\end{algorithmic}
\end{algorithm}

\begin{algorithm}[t]
\caption{Inference Procedure: Forward Direction}\label{alg:inference}

\textbf{Input:} Input $x$, training dataset $D = \{ (x_i, y_i) \}$, trained model $\Phi$, with neighborhood centers  $\mathcal{R}_X$ and $\mathcal{R}_Y$ and priors $\mathcal{D}_1$ and $\mathcal{D}_2$  \\
\textbf{Output:} A set $S$ of $n$ samples from the distribution $f(x)$
\begin{algorithmic}[1]
\State $S = \{ \}$
\State  $r_i^X = \argmin_{r \in \mathcal{R}_X} |r - x |$
\State Construct a dataset $D_x = \{ (x_i, y_i) | (x_i, y_i) \in D \text{ and } x_i \in \text{Neigh}(x) \}$ 

\While{$|S| < n $}
\State Sample  $y' \sim \text{proj}_y(D_x)$ 
\State  $r_j^Y = \argmin_{r \in \mathcal{R}_Y} |r - y' |$
\State Sample $z \sim \mathcal{D}_1^i$ 
\State $(\hat{y}, \hat{z_2}) =  \Phi(x,z,r^X, r^Y)$
\State $S = S \cup  \{ \hat{y} \}$
\EndWhile

\end{algorithmic}
\end{algorithm}

\section{Experiments}

We evaluate on three synthetic datasets. The first dataset, {\emph{Torus-to-circle I}}, is illustrated in Figure~\ref{fig:dataset1} and consists of a map from the torus to a circle. In the forward direction, (Figure~\ref{fig:dataset1}, center), each slice of the torus maps to a random point sampled uniformly from an interval of the circle centered on the projection of that slice. The second dataset, {\emph{Torus-to-circle II}} ( Figure~\ref{fig:dataset2}), similarly maps the torus to the circle. In the forward direction a slice of the torus with axis of revolution angle $\theta$ is projected to one of two points on the circle with angles $\theta/2$ or $\theta/2 + \pi$, with equal probability. The third dataset, {\emph{M\"obius-to-circle}}, is very similar to {\emph{Torus-to-circle I}}, with the torus replaced by a M\"obius band.  More details on each dataset can be found in Section \ref{appendix-datasets} of the Appendix. 

We compare BMNet to two other many-to-many models, Augmented CycleGAN (AugCGAN) \citep{almahairi2018} and Latent Normalizing Flows for Many-to-Many Cross-Domain Mappings (LNFMM)  \citep{mahajan2020latent}. Additionally, we include a pair of cGAN models \citep{mirza2014conditional} and a pair of cNF models \citep{winkler2019}. Note that in order to adapt cGANs and cNFs for many-to-many problems, we train one cGAN and cNF for the forward direction and one for the backward direction. We follow the evaluation procedure proposed in \cite{courts2021bundle}. In the forward direction, to evaluate the global distribution, we generate 5000 points and compare to 5000 points from the true distribution. To evaluate local distributions, we sample 15 points $U$ from $X$ and for each $x \in U$, use the model to generate 200 points in $Y$ and compare this to the true distribution defined by the dataset. Evaluation of the backward direction is analogous. To compare true and reconstructed distributions we use a range of different metrics, including the 1-Wasserstein metric, mean squared minimum distance (MSMD), maximum mean discrepancy (MMD), and KL divergence. To obtain confidence intervals, we train each model 5 times and report the mean with $95 \%$ confidence intervals.


\subsection{Results}

\begin{wrapfigure}{r}{0.5\linewidth}
    \vspace{-10mm}
    \centering
    \includegraphics[ scale=0.25]{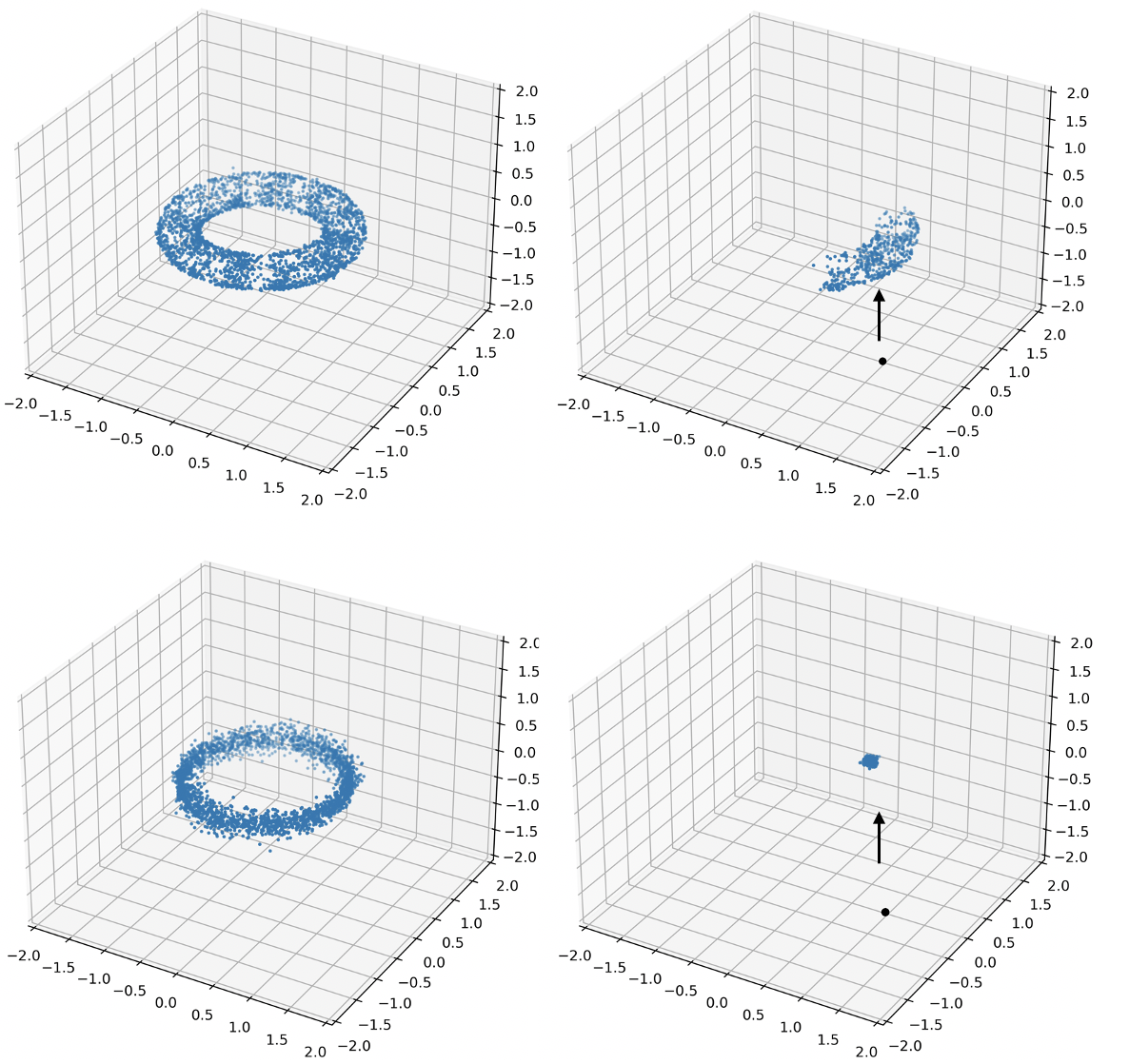}
   \caption{Global and local distributions generated by our model (top row) and an AugCGAN (bottom row) for Torus-to-circle I in the reverse direction.}
   \label{fig:sampled_torus_dataset1}
\end{wrapfigure}

As can be seen in Table ~\ref{tab:combined-results-global}, at the global level our model is the best performing model in terms of the Wasserstein-1 distance on most of the datasets. While AugCGAN performs well at the global level, as seen in Table ~\ref{tab:combined-results-local}, this many-to-many model does not learn the local distributions. The cNF model also generally performs well, particularly at the local level, though requires training a separate model in each direction and does not learn the discrete nature of the Torus-to-circle II dataset in the forward direction. This is depicted in Appendix \ref{appendix-visualizations}. 

To give a flavor of the difference in reconstructions for models explicitly designed to model many-to-many processes, the reader should consult Figure~\ref{fig:sampled_torus_dataset1} which depicts the global and local distributions of this dataset reconstructed by BMNet (top row) and an AugCGAN (bottom row). Locally, while the AugCGAN struggles to learn the full range of possible inputs to achieve a given $y$, our model is able to recreate a reasonable approximation of cylindrical sections of the torus. In general, we see this pattern repeated across datasets in both the forward and backward direction. This suggests to us that the structure of BMNet makes it more capable of handling the problem set forth in this short paper. Full results for each dataset, all models, all metrics, and further visualizations are available in Sections\ref{appendix-visualizations} and  \ref{Appendix-full-results} of the Appendix. 

\begin{table}[th]
\caption{The Wasserstein-1 ($\times 10^{-2}$) global metric on all datasets and models. }
\label{tab:combined-results-global}
\begin{center}
\begin{tabular}{r|rrrrr}
 &  \small{BMNet } & \small{AugCGAN } &\small{ LNFMM } & \small{cGAN }   & \small{cNF } 
 \\ \hline 
\scriptsize{Torus-to-circle I: Fwd} & \small{$1.91\pm 0.34$} & \small{$9.80\pm 0.88 $}& \small{$15.01\pm 5.55 $} & \small{$4.38\pm 1.61$} & \small{$\mathbf{1.28\pm 0.16}$} \\
\scriptsize{Torus-to-circle I: Rev}& \small{$\mathbf{4.38\pm 0.57}$}
& \small{$19.13\pm 3.51 $}& \small{$25.95\pm 3.16 $}
& \small{$28.96\pm 3.72$} & \small{$7.10\pm 0.58$}\\
\scriptsize{Torus-to-circle II: Fwd} & \small{$\mathbf{0.38\pm 0.10}$} 
& \small{$0.45\pm 0.05 $}& \small{$23.36\pm 5.94 $}
& \small{$6.29\pm 0.84$} & \small{$27.54\pm 5.49$} \\
\scriptsize{Torus-to-circle II: Rev} &  \small{$\mathbf{3.51\pm 0.17}$}
& \small{$9.48\pm 1.0 $}& \small{$20.54\pm 0.35 $}
& \small{$48.84\pm 12.01$} &  \small{$8.13\pm 0.70$}\\
\scriptsize{M\"obius-to-circle: Fwd } & \small{$1.85\pm 0.39$} 
& \small{$\mathbf{0.68\pm 0.11} $}& \small{$6.50\pm 2.62 $}
& \small{$8.49\pm 1.54$} & \small{$2.58\pm 0.59$}\\
\scriptsize{M\"obius-to-circle: Rev} &  \small{$\mathbf{2.33\pm 0.19}$} 
& \small{$2.37\pm 0.16 $}& \small{$23.89\pm 11.95 $}
& \small{$103.7\pm 4.1$} &  \small{$3.36\pm 0.05$}
\end{tabular}
\end{center}
\end{table}

\begin{table}[th]
\caption{The Wasserstein-1 ($\times 10^{-2}$) local metric on all datasets and models. }
\label{tab:combined-results-local}
\begin{center}
\begin{tabular}{r|rrrrr}
 &  \small{BMNet } & \small{AugCGAN } &\small{ LNFMM } & \small{cGAN }   & \small{cNF } 
 \\ \hline 
\scriptsize{Torus-to-circle I: Fwd} &  \small{$9.30\pm 1.26$} & \small{$119.9\pm 10.4 $}& \small{$20.31\pm 2.22 $} & \small{$20.33\pm 2.01$} & \small{$\mathbf{5.68\pm 0.72}$}\\
\scriptsize{Torus-to-circle I: Rev}& \small{$\mathbf{12.01\pm 0.95}$}
& \small{$123.6\pm 9.9 $}& \small{$46.23\pm 0.71 $}
& \small{$37.48\pm 1.80$}& \small{$12.43\pm 0.28$}\\
\scriptsize{Torus-to-circle II: Fwd } & \small{$16.05\pm 4.38$} 
& \small{$127.1\pm 2.8 $}& \small{$31.71\pm 2.29 $}
& \small{$\mathbf{9.37\pm 1.01}$}& \small{$32.86 \pm 2.26$}\\
\scriptsize{Torus-to-circle II: Rev} &  \small{$\mathbf{4.49\pm 0.54}$}
& \small{$116.1\pm 13.7 $}& \small{$23.58\pm 0.15 $}
& \small{$56.39\pm 5.73$} & \small{$10.36\pm 0.43$}\\
\scriptsize{M\"obius-to-circle: Fwd } & \small{$9.48\pm 0.98$} 
& \small{$87.3\pm 13.41 $}& \small{$16.67\pm 2.08 $}
& \small{$42.63\pm 3.71$}&  \small{$\mathbf{6.08 \pm 0.93}$}\\
\scriptsize{M\"obius-to-circle: Rev} &  \small{$8.77\pm 0.79$}
& \small{$94.32\pm 14.2 $}& \small{$42.78\pm 2.03 $}
& \small{$117.17\pm 1.84$}& 
\small{$\mathbf{8.60\pm 0.38}$}
\end{tabular}
\end{center}
\end{table}

\section{Conclusion}

Many-to-many processes are common in nature. While there are a range of deep learning-based frameworks that can be used to solve simple tasks related to these processes, network architectures for more comprehensive modeling have until now remained limited. Guided by the concept of a bundle morphism, in this paper we introduce the first model architecture explicitly designed to capture the more nuanced aspects of many-to-many processes, providing the capability to model not only the distribution as a whole, but also the distributions of both the image of points and their fibers. 


\subsubsection*{Acknowledgments}
This research was supported by the Mathematics for Artificial Reasoning in Science (MARS) initiative via the Laboratory Directed Research and Development (LDRD) investments at Pacific Northwest National Laboratory (PNNL). PNNL is a multi-program national laboratory operated for the U.S. Department of Energy (DOE) by Battelle Memorial Institute under Contract No. DE-AC05-76RL0-1830.

\bibliography{references}
\bibliographystyle{iclr2022_workshop}

\appendix
\section{Appendix}

\subsection{Related Work}

As discussed above, this work builds on \citet{courts2021bundle} where a fiber bundle-inspired deep learning framework was used to address the problem of modeling the fibers in many-to-one machine learning tasks. Our work differs from that one in that (i) the problem (modeling both fibers and images in many-to-many maps) is distinct from the problem that is considered in \citet{courts2021bundle}, (ii) the inspiration for BMNet's architecture is the concept of a bundle morphism, while the inspiration for BundleNet was an individual fiber bundle, and (iii) BMNet's training routine and architecture are distinct from BundleNet.

Beyond BundleNet, fiber bundles (and more specifically vector bundles) have recently begun to see more use as a tool in data science. \cite{scoccola2021approximate} for example, seek to develop the theory required to use vector bundles in a rigorous way in data science. Recent applications of vector bundles include: cryo-electron microscopy \citep{ye2017cohomology,gao2021geometry} and computer graphics \citep{knoppel2016complex}. The present work differs from those listed above in that we only use the concept of a bundle morphism to guide the problem framework and model architecture for learning many-to-many maps.

More broadly, deep learning-based generative models have recently made great strides forward, both in terms of the complexity of distributions that can be modeled and the fidelity of reconstructions. While the majority of these approaches seek to model a single distribution, there have also been a significant number of efforts to develop conditional generative approaches which are capable of modeling multiple distributions conditioned on additional input. These include conditional variational autoencoders \citep{sohn2015learning} and conditional generative adversarial networks \citep{mirza2014conditional}. One of the results of this paper was to show that like the many-to-one fiber modeling problem introduced in \citep{courts2021bundle}, the many-to-many fiber and image modeling problem cannot be satisfactorily addressed with existing architectures. Recent topological approaches to evaluating generative model performance includes \citep{ zhou2020evaluating}.

\subsection{Limitations}

Because it fit well within the training routine we wanted to use, we chose to use an architecture that is invertible after conditioning (as in \citet{courts2021bundle}). Of course, in practice bundle morphisms need not be invertible (though the local trivializations that are built into our model are) and it is likely that this constraint will hurt performance on certain types of datasets. In the future it would be worth revisiting the training procedure to understand how to include non-invertible components into the architecture. 

Beyond this, the major limitation of this work is that the experiments that we describe are limited to synthetic datasets. We felt that this was a reasonable choice based on space limitations and the novelty of the problem, but future work should identify and test against real-world many-to-many datasets. One challenge related to the former is that most existing benchmark ML datasets are chosen to be `nice' and thus there is a bias against one-to-many type phenomena.

\subsection{The Analogy Between a Bundle Morphism Network and a True Bundle Morphism}

We caution that while BMNet is inspired by the idea of a bundle morphism, we did not strictly constrain the model to this definition. In this section we briefly lay out the analogy between components of the model architecture and a bundle morphism. To fix notation assume that $\mathbf{E_1} = (E_1,B_1,Z_1,\pi_1)$ and $\mathbf{E_2} = (E_2,B_2,Z_2,\pi_2)$ are two fiber bundles where $E_i$ is the total space, $B_i$ is the base space, $Z_i$ is the fiber, and $\pi_i$ is the projection operator for $i \in \{1,2\}$. Also let $F: E_1 \rightarrow E_2$ be a bundle morphism that descends to a continuous map $f:B_1 \rightarrow B_2$ making the required diagram commute. On the other hand, let $\Phi: X \times Z_1 \times \mathcal{R}_X \times \mathcal{R}_Y \rightarrow Y \times Z_2$ be an instance of BMNet.
\begin{itemize}
    \item $B_1$ and $B_2$ correspond to $X$ and $Y$ respectively. 
    \item $Z_1$ and $Z_2$ in the fiber bundle set-up correspond to $Z_1$ and $Z_2$ in the description of BMNet. Note the distributions associated with $Z_1$ and $Z_2$ in BMNet are critical to the model, whereas probability distributions on the fibers of fiber bundles are not commonly considered in topology or geometry. This may be an interesting area for further investigation.
    \item The representatives $\mathcal{R}_X = \{r_1^X, \dots, r_{n_1}^X\}$ and $\mathcal{R}_Y = \{r_1^Y, \dots, r_{n_2}^Y\}$ correspond to the local trivialization neighborhoods in $B_1$ and $B_2$. 
    \item $\Phi_{i,j}$ corresponds to $\phi_j \circ F \circ \phi_i^{-1}$, where $\phi_i: \pi^{-1}_1(U_i) \rightarrow U_i \times F_1$ and $\phi_j: \pi^{-1}_2(U_j) \rightarrow U_j \times F_2$ are local trivializations between neighborhoods $\pi_1^{-1}(U_i) \subseteq E_1$ and $\pi_2^{-1}(U_j) \subseteq E_2$ respectively.
\end{itemize}

\subsection{Datasets}
\label{appendix-datasets}
 
\begin{figure}[h]
    \centering
    \includegraphics[ scale=0.2]{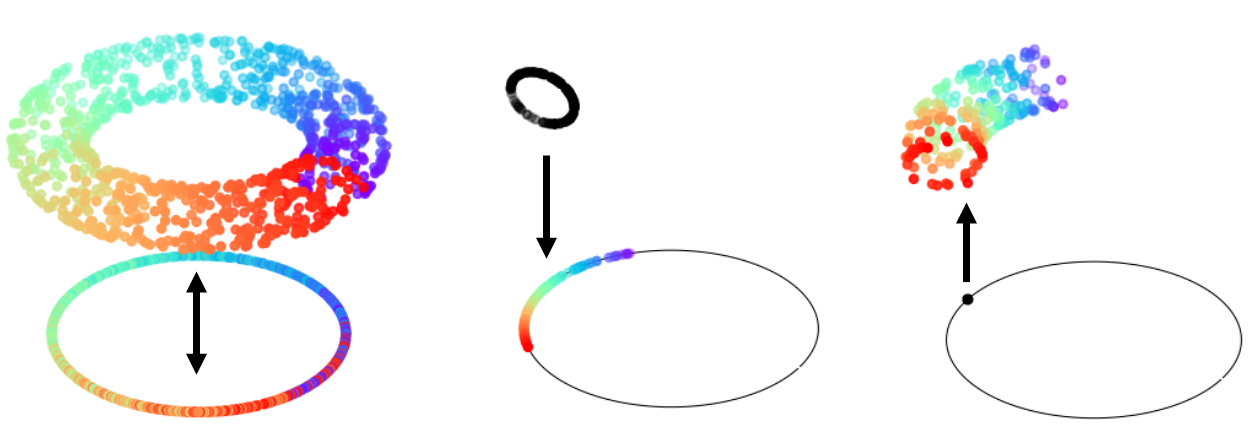}
  \caption{The {\emph{Torus-to-circle I}} many-to-many dataset (left). Each torus slice maps to a uniformly sampled random point on an interval of the circle (center).}
  \label{fig:dataset1}
\end{figure}

{\textbf{Torus-to-circle I:}} Here we describe the synthetic datasets used in more detail. For the Torus-to-circle I dataset, given radius $r$ and radius $R$, the torus representation is: 
\begin{equation*}
T = \{((R + r \:  \cos \, \theta) \:  \cos \, \phi, 
(R + r \:  \cos \, \theta) \:  \sin \, \phi,
r \: \sin \, \theta) 
: 0 \leq \theta, \phi \leq 2 \pi \}
\end{equation*}
We sample from $T$ with $R = 1$ and $r = 0.25$ by randomly sampling $\theta$ and $\phi$. We pair this with a point from the circle by randomly sampling a perturbation $\alpha$ to add to $\phi$. : 
\begin{equation*}
C = \{( \cos(\phi + \alpha), \sin(\phi + \alpha), 
: 0 \leq  \phi \leq 2 \pi , \frac{-\pi}{4} \leq  \alpha \leq \frac{\pi}{4} \}
\end{equation*}
This dataset is visualized in Figure \ref{fig:dataset1}.

\begin{figure}[h]
    \centering
    \includegraphics[ scale=0.25]{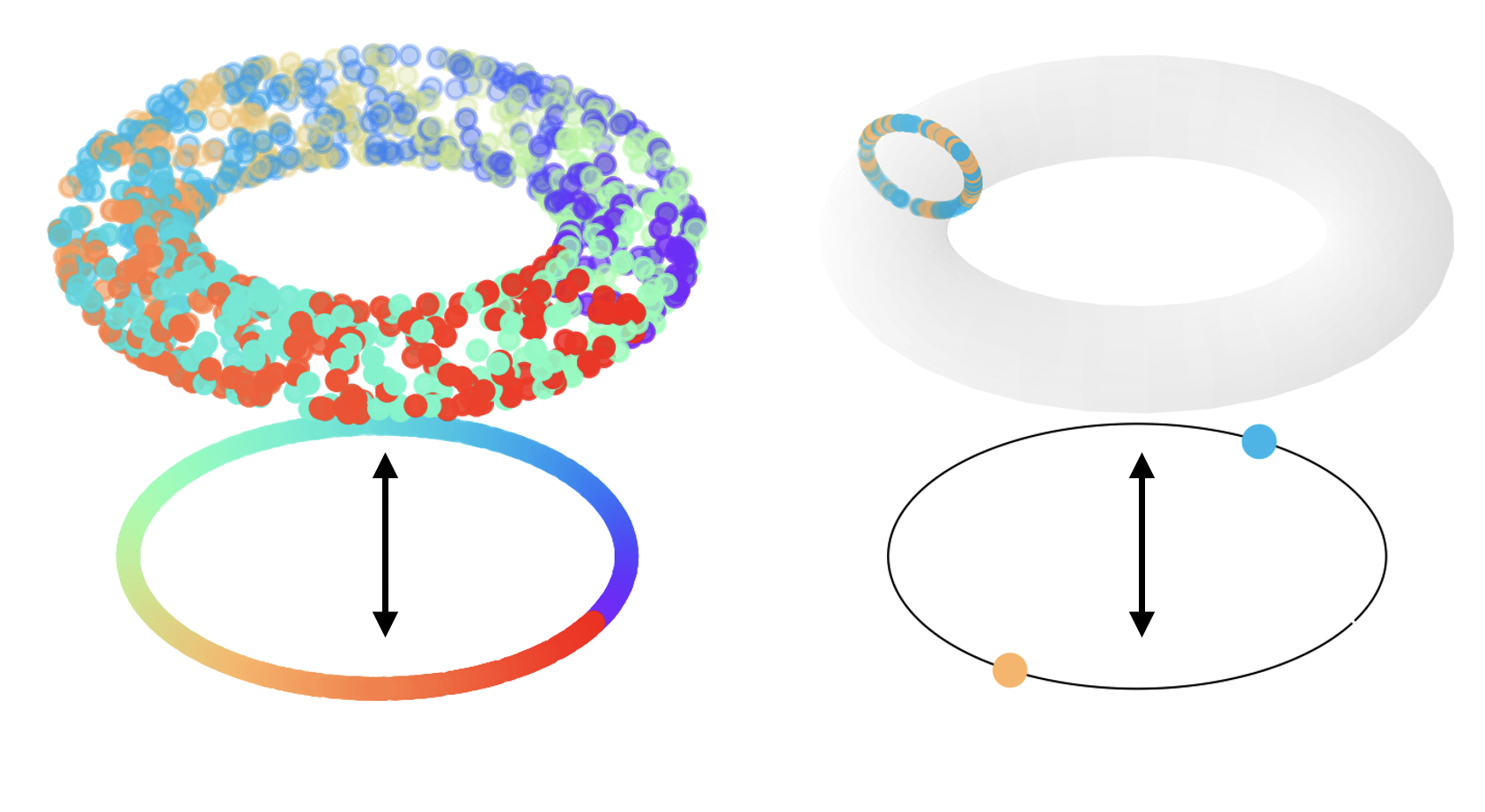}
  \caption{The {\emph{Torus-to-circle II}} dataset. A point at angle $\theta$ of the torus maps to one of two points on the circle, $\theta/2$ and $\theta/2 + \pi$ (right) with equal probability.}
  \label{fig:dataset2}
\end{figure}

{\textbf{Torus-to-circle II:}} Generation of a point from the Torus-to-circle II dataset begins by sampling a point from $T$ as above and then pairing that with one of two points from $C$, either $(\cos(\theta/2),\sin(\theta/2))$ or $(\cos(\theta/2 + \pi),\sin(\theta/2 + \pi))$ with equal probability. This dataset is depicted in  Figure~\ref{fig:dataset2}.




{\textbf{M\"obius-to-circle:}} The M\"obius-to-circle dataset is given by: 
\begin{equation*}
M = \{(R \, \cos \theta - s \, \cos \frac{\theta}{2}  \cos \theta, 
R  \, \sin \theta  - s \, \cos \frac{\theta}{2}  \sin \theta,
s \, \sin \frac{\theta}{2} )
: 0 \leq \theta \leq 2 \pi , -r \leq s \leq r \}
\end{equation*}

We sample from $M$ with $R = 1$ and $r = 0.25$ by randomly sampling $\theta$ and $s$. We pair this with a point from the circle as in  Torus-to-Circle I.

\subsection{Training Details}

Each model was trained on 5,000 training examples on a single GPU for 2,000 epochs. An initial learning rate of $10^{-4}$ was used, and was reduced by a factor of 10 at 1000 and 1500 epochs.  

Experimentally, we have found that the most effective prior to impose on $Z_1$ and $Z_2$ was a uniform circular prior. During training, we update the parameters of the prior distributions based on the learned $z_1$ and $z_2$. Additionally, we enforce a regularization term on $z_1$ and $z_2$ for model stability during training. 

\subsection{Visualizing reconstructions}
\label{appendix-visualizations}

\begin{figure}[h!]
    \centering
    \includegraphics[ scale=0.35]{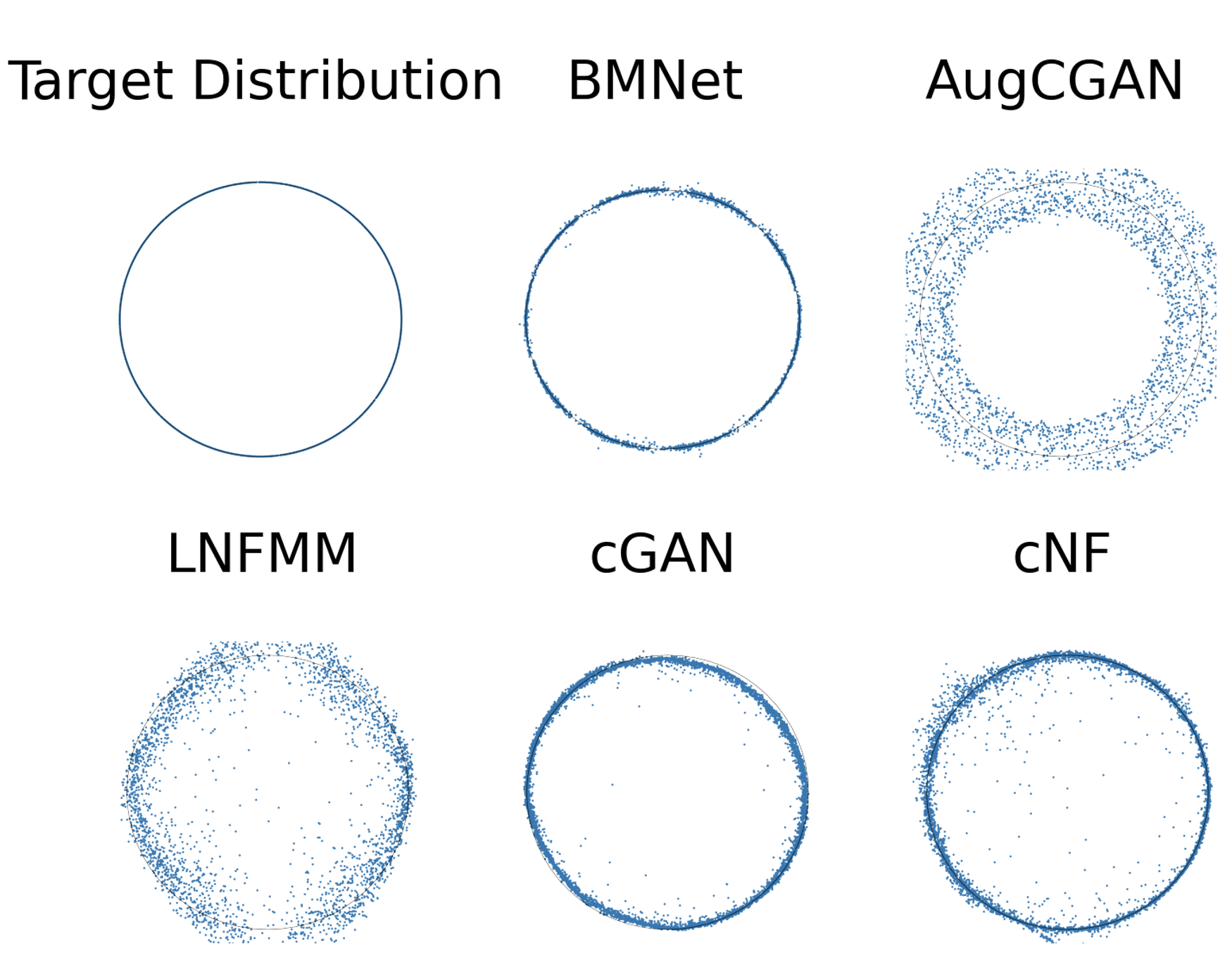}
   \caption{Global distributions generated by the different models in the forward direction on Torus-to-circle I.}
   \label{fig:sampled_circle_dataset1}
\end{figure}

\begin{figure}[h!]
    \centering
    \includegraphics[ scale=0.35]{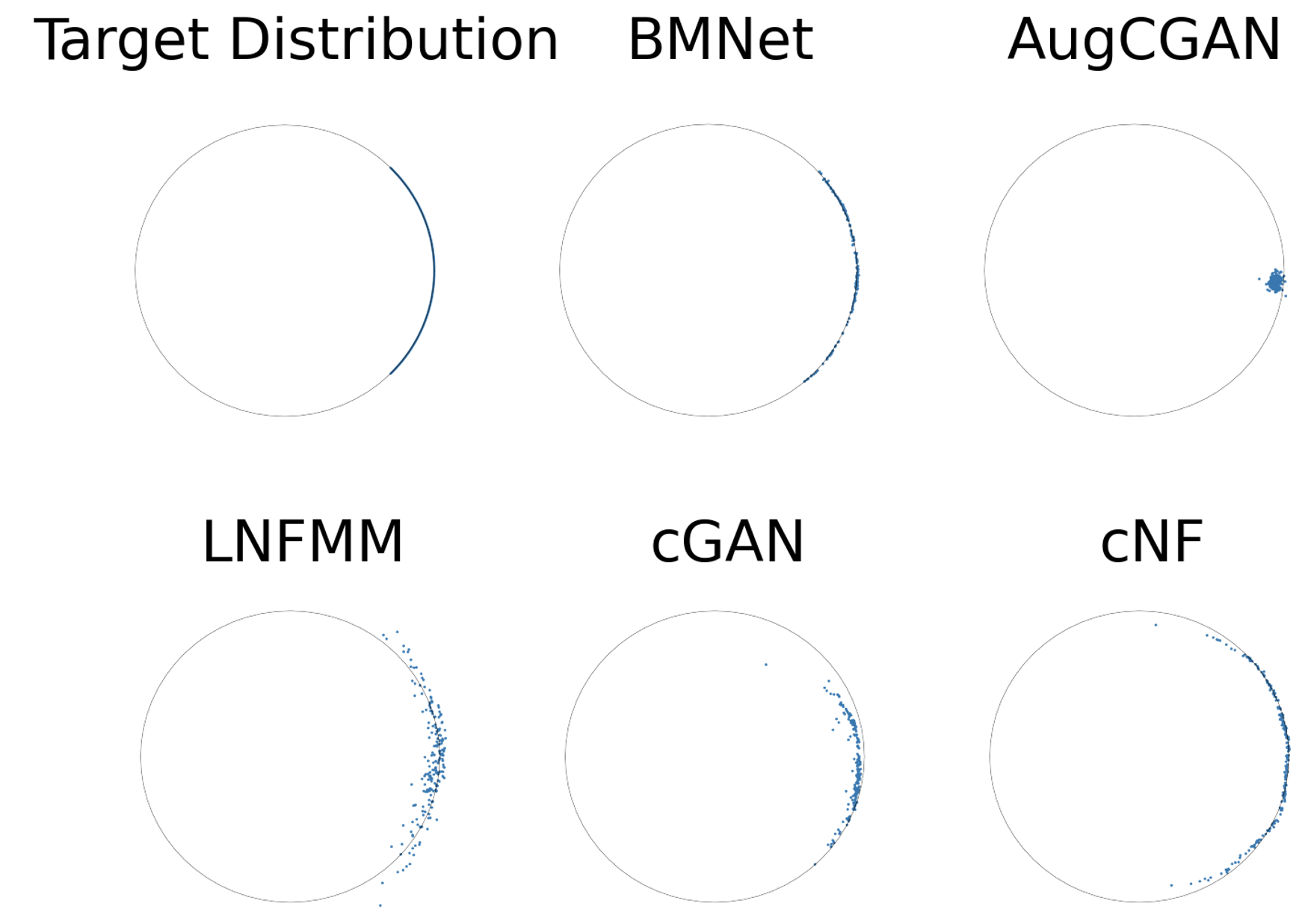}
   \caption{Local distributions generated by the different models in the forward direction on Torus-to-circle I.}
   \label{fig:sampled_local_circle_dataset1}
\end{figure}

\begin{figure}[h!]
    \centering
    \includegraphics[ scale=0.35]{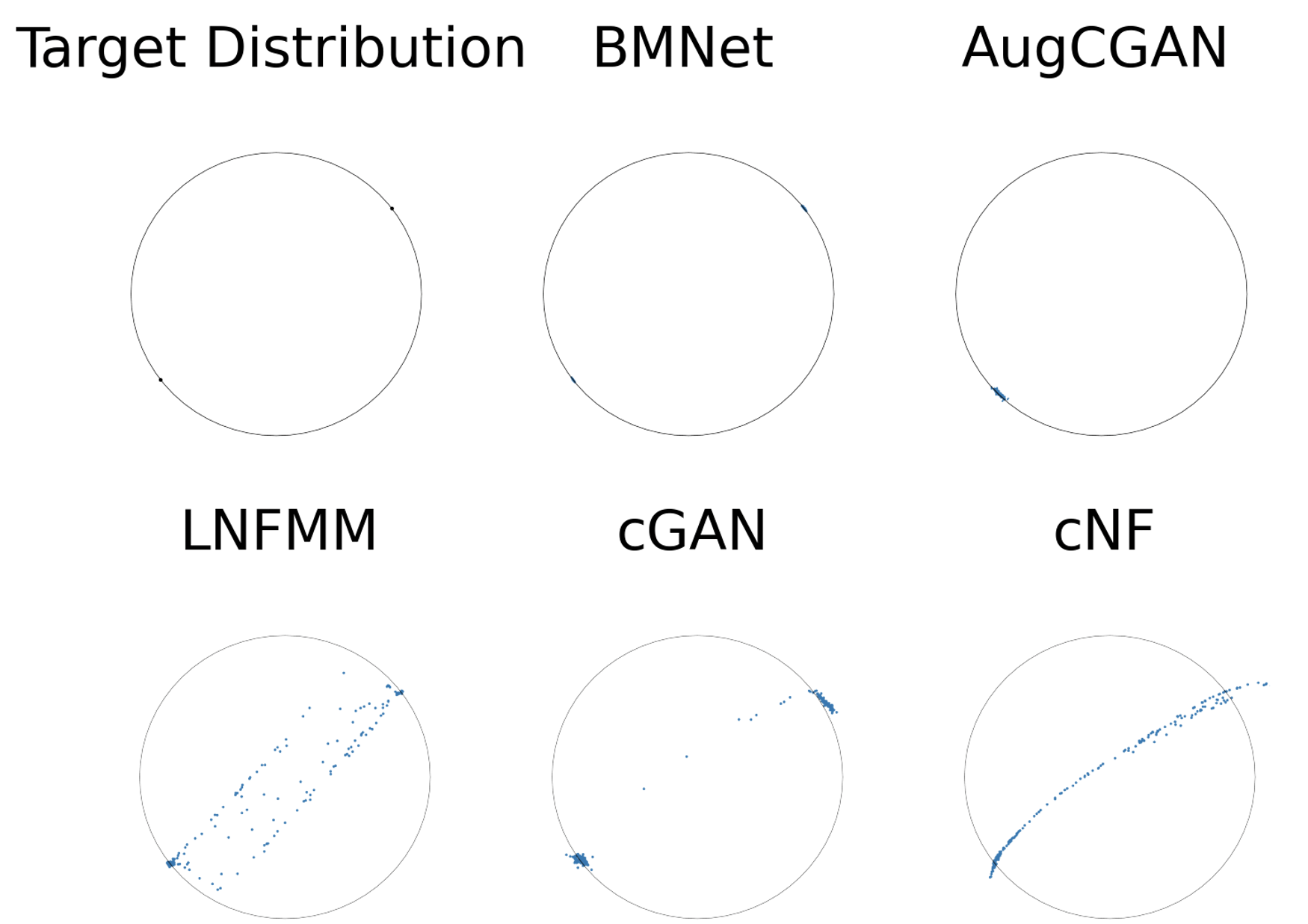}
   \caption{Local distributions generated by the different models in the forward direction on Torus-to-circle II.}
   \label{fig:sampled_local_circle_dataset2}
\end{figure}

\begin{figure}[h!]
    \centering
    \includegraphics[ scale=0.35]{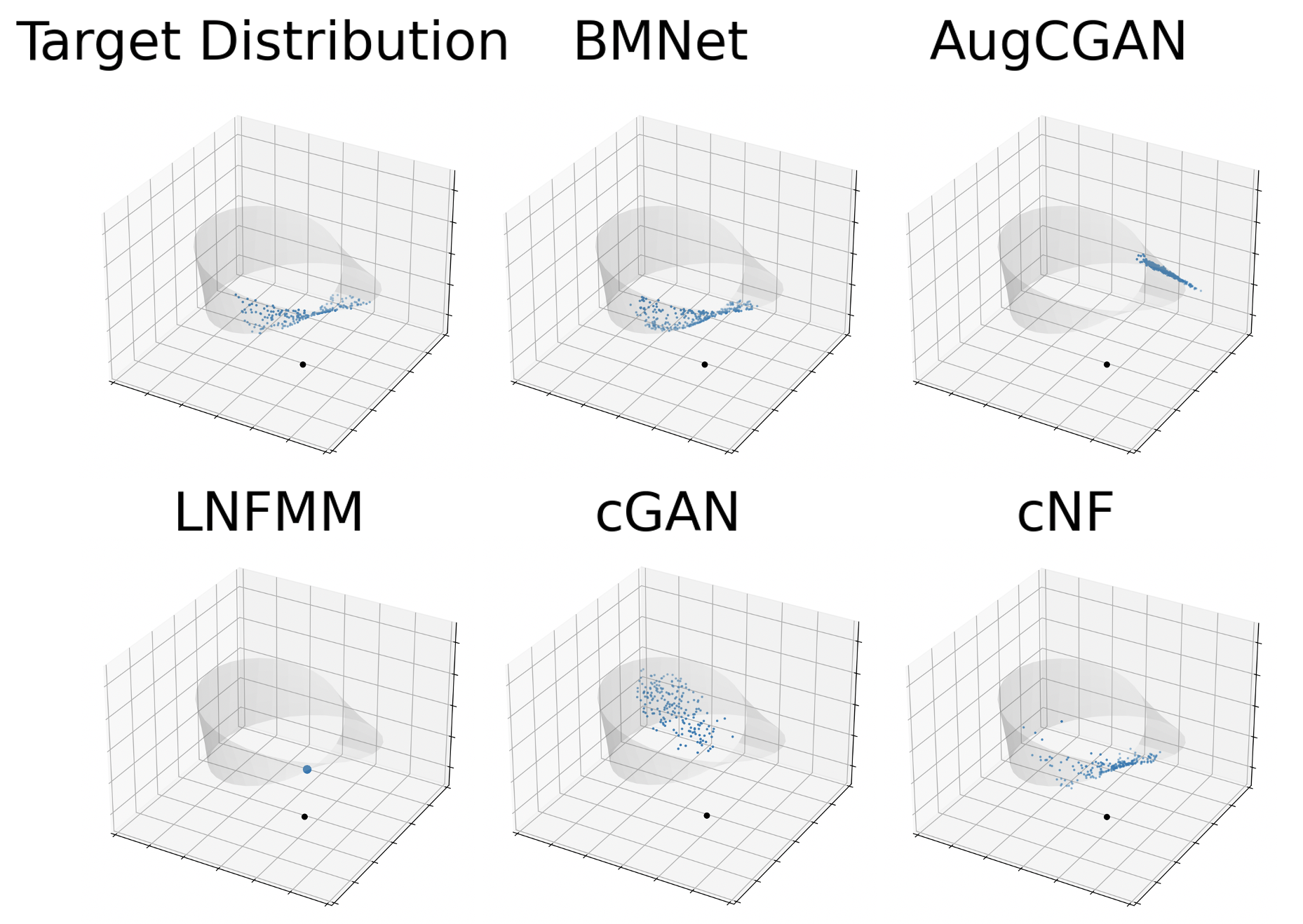}
   \caption{Local distributions generated by the different models in the reverse direction on M\"obius-to-circle.}
   \label{fig:sampled_local_moeb_dataset3}
\end{figure}

In this section we provide visualizations of the reconstructions of BMNet and the other models in Figures \ref{fig:sampled_circle_dataset1} through \ref{fig:sampled_local_moeb_dataset3}.

\subsection{Full Results}
\label{Appendix-full-results}

We provide full results with all metrics for both models on all three datasets in Tables \ref{table:torus-to-circle-I-fwd} to \ref{table:mob-to-circle-bwd}.

\begin{table}[th!]
\caption{All metrics Torus-to-circle I: forward direction (global)}
\label{table:torus-to-circle-I-fwd}
\begin{center}
\begin{tabular}{r|rrrrr}
 & \small{BMNet } & \small{AugCGAN } &\small{ LNFMM } & \small{cGAN }   & \small{cNF }
 \\ \hline 
\small{MSMD ($\times 10^{-4}$)} &  \small{$\mathbf{3.08\pm 2.56}$} &  \small{$189.8\pm 24.9 $}& \small{$661.2\pm 428.6 $} & \small{$44.80\pm 20.97$} & \small{$17.24\pm 7.06$}\\

\small{MMD ($\times 10^{-3}$)}& \small{$0.54\pm 0.29$} & \small{$0.82\pm 0.13 $}& \small{$6.27\pm 3.0 $} & \small{$2.35\pm 1.22$} &  \small{$\mathbf{0.49\pm 0.13}$}\\
\small{KL-fwd } & \small{$\mathbf{0.93\pm 0.10}$} &  \small{$4.42\pm 0.16 $}& \small{$4.21\pm 0.36 $} & \small{$3.93\pm 0.49$} & \small{$1.27\pm 0.07$}\\

\small{KL-bwd } &\small{$\mathbf{3.52\pm 0.21}$}& \small{$6.97\pm 0.12 $}& \small{$6.79\pm 0.44 $} & \small{$6.71\pm 0.54$} & \small{$3.82\pm 0.10$}\\

\small{$\mathcal W_1$ ($\times 10^{-2}$)}&  \small{$1.91\pm 0.34$} & \small{$9.80\pm 0.88 $}& \small{$15.01\pm 5.55 $} & \small{$4.38\pm 1.61$} & \small{$\mathbf{1.28\pm 0.16}$} \\

\small{$\mathcal W_2$ ($\times 10^{-2}$)} & \small{$\mathbf{0.11\pm 0.06}$}& \small{$0.97\pm 0.13 $}& \small{$2.62\pm 1.42 $} & \small{$0.56\pm 0.27$} & \small{$0.12\pm 0.09$}  
\end{tabular}
\end{center}
\end{table}

\begin{table}[th!]
\caption{All metrics Torus-to-circle I: forward direction (local)}
\label{table:torus-to-circle-I-fwd-local}
\begin{center}
\begin{tabular}{r|rrrrr}
 & \small{BMNet} & \small{AugCGAN } &\small{ LNFMM } & \small{cGAN }   & \small{cNF } 
 \\ \hline 
\small{MSMD ($\times 10^{-4}$)} &  \small{$\mathbf{13.98\pm 4.09}$} 
& \small{$7127 \pm 1230  $}& \small{$400.1\pm 142.1 $}
& \small{$49.27\pm 13.87$}
& \small{$75.82\pm 32.74$}   \\
\small{MMD ($\times 10^{-3}$)}&  \small{$14.48\pm 3.31$} 
& \small{$612.4\pm 57.6 $}& \small{$37.33\pm 5.50 $}
& \small{$64.29\pm 8.69$} & \small{$\mathbf{11.09\pm 2.45}$} \\
\small{KL-fwd } &  \small{$1.57\pm 0.29$} & \small{$9.87\pm 0.46 $}& \small{$3.66\pm 0.37 $} & \small{$3.50\pm 0.33$} & \small{$\mathbf{0.71\pm 0.10}$} 
\\
\small{KL-bwd } & \small{$3.64\pm 0.42$}& \small{$12.05\pm 0.32 $}& \small{$5.90\pm 0.30 $} & \small{$5.92\pm 0.35$} & \small{$\mathbf{2.38\pm 0.16}$} \\
\small{$\mathcal W_1$ ($\times 10^{-2}$)}& \small{$9.30\pm 1.26$} & \small{$119.9\pm 10.4 $}& \small{$20.31\pm 2.22 $} & \small{$20.33\pm 2.01$} & \small{$\mathbf{5.68\pm 0.72}$} \\
\small{$\mathcal W_2$ ($\times 10^{-2}$)} & \small{$1.37\pm 0.32$} & \small{$90.27\pm 11.1 $}& \small{$4.78\pm 0.95 $} & \small{$4.33\pm 0.62$} & \small{$\mathbf{0.94\pm 0.94}$}  
\end{tabular}
\end{center}
\end{table}

\begin{table}[th!]
\caption{All metrics Torus-to-circle I: reverse direction (global) }
\label{table:torus-to-circle-I-bwd}
\begin{center}
\begin{tabular}{r|rrrrr}
 &  \small{BMNet } & \small{AugCGAN } &\small{ LNFMM } & \small{cGAN }   & \small{cNF }  \\ \hline 
\small{MSMD ($\times 10^{-4}$) } & \small{$\mathbf{7.39 \pm 0.58}$} & \small{$272.4\pm 186.2 $}& \small{$193.2\pm 24.9 $} &  \small{$257.4\pm 88.5$}  & \small{$87.86\pm 15.23$}\\
\small{MMD ($\times 10^{-3}$)}& \small{$1.59\pm 0.78$}
& \small{$8.93\pm 4.91 $}& \small{$16.51\pm 9.64 $}
& \small{$29.46\pm 10.50$}  & \small{$\mathbf{0.38\pm 0.13}$}\\
\small{KL-fwd } & \small{$\mathbf{0.18\pm 0.07}$} 
& \small{$4.41\pm 0.72 $}& \small{$13.85\pm 3.16 $}
& \small{$6.28\pm 0.88$}  & \small{$1.04\pm 0.19$}\\
\small{KL-bwd } &\small{$\mathbf{0.64\pm 0.16}$} 
& \small{$5.71\pm 0.67 $}& \small{$7.65\pm 0.12 $}
& \small{$6.62\pm 0.55$} 
& \small{$2.05\pm 0.13$}\\
\small{$\mathcal W_1$ ($\times 10^{-2}$)}& \small{$\mathbf{4.38\pm 0.57}$}
& \small{$19.13\pm 3.51 $}& \small{$25.95\pm 3.16 $}
& \small{$28.96\pm 3.72$} & \small{$7.10\pm 0.58$}\\
\small{$\mathcal W_2$ ($\times 10^{-2}$)}& \small{$\mathbf{0.32\pm 0.09}$} 
& \small{$2.46\pm 1.01 $}& \small{$4.67\pm 1.33 $} & \small{$6.78\pm 1.73$} & \small{$12.43\pm 0.28$} 
\end{tabular}
\end{center}
\end{table}

\begin{table}[th!]
\caption{All metrics Torus-to-circle I: reverse direction (local) }
\label{table:torus-to-circle-I-bwd}
\begin{center}
\begin{tabular}{r|rrrrr}
 &  \small{BMNet } & \small{AugCGAN } &\small{ LNFMM } & \small{cGAN }   & \small{cNF } 
 \\ \hline 
\small{MSMD ($\times 10^{-4}$) } & \small{$\mathbf{46.28\pm 2.60}$} 
& \small{$6422 \pm 1005 $}& \small{$243.1\pm 28.2 $}
& \small{$258.0\pm 44.4$} & \small{$153.0\pm 9.4$} \\
\small{MMD ($\times 10^{-3}$)}& \small{$13.00\pm 2.70$} 
& \small{$554.6\pm 55.9 $}& \small{$127.5\pm 9.0 $}
& \small{$113.6\pm 12.2$} & \small{$\mathbf{4.67\pm 0.65}$} \\
\small{KL-fwd } &  \small{$0.90\pm 0.21$}
& \small{$10.82\pm 0.71 $}& \small{$26.4\pm 1.59 $}
& \small{$5.51\pm 0.29$} & \small{$\mathbf{0.46\pm 0.04}$} \\
\small{KL-bwd }  & \small{$\mathbf{0.89\pm 0.20}$} 
& \small{$9.13\pm 0.41 $}& \small{$6.72\pm 0.04 $}
& \small{$4.74\pm 0.14$} & \small{$1.40\pm 0.05$}\\
\small{$\mathcal W_1$ ($\times 10^{-2}$)}&  \small{$\mathbf{12.01\pm 0.95}$}
& \small{$123.6\pm 9.9 $}& \small{$46.23\pm 0.71 $}
& \small{$37.48\pm 1.80$}& \small{$12.43\pm 0.28$}\\
\small{$\mathcal W_2$ ($\times 10^{-2}$)}&\small{$1.40\pm 0.22$} 
& \small{$95.18\pm 10.78 $}& \small{$13.64\pm 0.38 $}
& \small{$9.72\pm 0.90$} & \small{$\mathbf{1.06\pm 0.06}$} 
\end{tabular}
\end{center}
\end{table}

\begin{table}[th!]
\caption{All metrics Torus-to-circle II: forward direction (global)}
\label{table:torus-to-circle-II-fwd}
\begin{center}
\begin{tabular}{r|rrrrr}
 &  \small{BMNet } & \small{AugCGAN } &\small{ LNFMM } & \small{cGAN }   & \small{cNF } 
 \\ \hline 
\small{MSMD ($\times 10^{-4}$) } & \small{$\mathbf{0.09\pm 0.06}$} 
& \small{$0.94\pm 0.13 $}& \small{$1397\pm 377 $}
& \small{$251.0\pm 27.7$} &  \small{$1571\pm 522$}\\
\small{MMD ($\times 10^{-3}$) }& \small{$\mathbf{0.17\pm 0.07}$} 
& \small{$0.26\pm 0.08 $}& \small{$26.02\pm 8.57 $}
& \small{$1.83\pm 0.41$} & \small{${33.90\pm 11.76}$}\\
\small{KL-fwd } & \small{$\mathbf{0.32\pm 0.16}$} 
& \small{$1.45\pm 0.17 $}& \small{$4.39\pm 0.7 $}
& \small{$3.53\pm 0.43$} & \small{$5.04\pm 0.53$} \\
\small{KL-bwd } &\small{$\mathbf{2.27\pm 0.38}$} 
& \small{$3.9\pm 0.14 $}& \small{$6.12\pm 0.87 $}
& \small{$6.11\pm 0.53$} &  \small{$7.01\pm 0.35$} \\
\small{$\mathcal W_1$ ($\times 10^{-2}$) }& \small{$\mathbf{0.38\pm 0.10}$} 
& \small{$0.45\pm 0.05 $}& \small{$23.36\pm 5.94 $}
& \small{$6.29\pm 0.84$} & \small{$27.54\pm 5.49$} \\
\small{$\mathcal W_2$ ($\times 10^{-2}$) }& \small{$\mathbf{0.04\pm 0.01}$} 
& \small{$0.06\pm 0.02 $}& \small{$6.28\pm 1.28 $}
& \small{$1.39\pm 0.14$} & \small{$6.08\pm 1.47$} 
\end{tabular}
\end{center}
\end{table}

\begin{table}[th!]
\caption{All metrics Torus-to-circle II: forward direction (local)}
\label{table:torus-to-circle-II-fwd}
\begin{center}
\begin{tabular}{r|rrrrr}
 &  \small{BMNet } & \small{AugCGAN } &\small{ LNFMM } & \small{cGAN }   & \small{cNF } 
 \\ \hline 
\small{MSMD ($\times 10^{-4}$) } &  \small{$\mathbf{3.70\pm 0.91}$} 
& \small{$8401 \pm 1411 $}& \small{$1831\pm 139 $}
& \small{$328.2\pm 72.6$} & \small{$1942\pm 195$}\\
\small{MMD ($\times 10^{-3}$) }&  \small{${38.49\pm 11.42}$}
& \small{$567.8\pm 38.8 $}& \small{$60.31\pm 7.57 $}
& \small{$\mathbf{9.37\pm 2.70}$} & \small{$65.62\pm 9.80$}\\
\small{KL-fwd }  & \small{$5.53\pm 0.18$}
& \small{$11.97\pm 0.47 $}& \small{$\mathbf{5.00\pm 0.11}$}
& \small{$5.28\pm 0.14$}& \small{$6.26\pm 0.16$}\\
\small{KL-bwd }  & \small{$\text{NA}$} 
& \small{$\text{NA}$}& \small{$\text{NA}$}
& \small{$\text{NA}$}& \small{$\text{NA}$}\\
\small{$\mathcal W_1$ ($\times 10^{-2}$) }& \small{$16.05\pm 4.38$} 
& \small{$127.09\pm 2.76 $}& \small{$31.71\pm 2.29 $}
& \small{$\mathbf{9.37\pm 1.01}$}& \small{$32.86 \pm 2.26$}\\
\small{$\mathcal W_2$ ($\times 10^{-2}$) } & \small{$12.75\pm 3.97$}
& \small{$98.67\pm 0.19 $}& \small{$12.62\pm 1.41 $}
& \small{$\mathbf{2.18\pm 0.38}$} & \small{$11.25\pm 1.02$} 
\end{tabular}
\end{center}
\end{table}

\begin{table}[th!]
\caption{All metrics Torus-to-circle II: reverse direction (global)}
\label{table:torus-to-circle-II-bwd}
\begin{center}
\begin{tabular}{r|rrrrr}
 &  \small{BMNet } & \small{AugCGAN } &\small{ LNFMM } & \small{cGAN }   & \small{cNF } 
 \\ \hline 
\small{MSMD ($\times 10^{-4}$) } & \small{$\mathbf{6.21 \pm 0.26 }$}
& \small{$160.2\pm 40.5 $}& \small{$419.4\pm 26.9 $}
& \small{$1154\pm 756$} &  \small{$113.6\pm 19.6$}\\
\small{MMD ($\times 10^{-3}$) }& \small{$0.98\pm 0.39$} 
& \small{$0.91\pm 0.25 $}& \small{$3.04\pm 0.13 $}
& \small{$49.84\pm 17.35$} &  \small{$\mathbf{0.45\pm 0.05}$}\\
\small{KL-fwd } & \small{$\mathbf{0.12\pm 0.05}$} 
& \small{$1.53\pm 0.15 $}& \small{$16.6\pm 0.32 $}
& \small{$7.16\pm 0.98$} & \small{$1.55\pm 0.22$}\\
\small{KL-bwd } &\small{$\mathbf{0.28\pm 0.08}$} 
& \small{$2.49\pm 0.09 $}& \small{$7.51\pm 0.07 $}
& \small{$7.04\pm 0.66$} &  \small{$2.40\pm 0.17$}\\
\small{$\mathcal W_1$ ($\times 10^{-2}$) }& \small{$\mathbf{3.51\pm 0.17}$}
& \small{$9.48\pm 1.0 $}& \small{$20.54\pm 0.35 $}
& \small{$48.84\pm 12.01$} &  \small{$8.13\pm 0.70$}\\
\small{$\mathcal W_2$ ($\times 10^{-2}$) }& \small{$0.21\pm 0.04$} &
\small{$0.24\pm 0.0 $}& \small{$2.61\pm 0.11 $} &
\small{$19.89\pm 7.42$} & \small{$\mathbf{0.08\pm 0.28}$} 
\end{tabular}
\end{center}
\end{table}

\begin{table}[th!]
\caption{All metrics Torus-to-circle II: reverse direction (local)}
\label{table:torus-to-circle-II-bwd}
\begin{center}
\begin{tabular}{r|rrrrr}
 &  \small{BMNet } & \small{AugCGAN } &\small{ LNFMM } & \small{cGAN }   & \small{cNF } 
 \\ \hline 
\small{MSMD ($\times 10^{-4}$) } & \small{$\mathbf{35.43 \pm 17.46}$} 
& \small{$1457\pm 2375 $}& \small{$403.7\pm 26.6 $}
& \small{$2671\pm 511$} & \small{$192.6\pm 18.3$}\\
\small{MMD ($\times 10^{-3}$) }&\small{$6.70\pm 1.30$} 
& \small{$568.0\pm 59.2 $}& \small{$21.68\pm 1.85 $}
& \small{$329.6\pm 49.8$} & \small{$\mathbf{3.94\pm 0.61}$}\\
\small{KL-fwd }  & \small{$\mathbf{2.52\pm 0.24}$} 
& \small{$9.83\pm 0.57 $}& \small{$33.94\pm 0.95 $}
& \small{$8.10\pm 0.35$} & \small{$2.66\pm 0.08$}\\
\small{KL-bwd } & \small{$\mathbf{5.64\pm 0.20}$}
& \small{$17.36\pm 0.65 $}& \small{$14.17\pm 0.08 $}
& \small{$14.84\pm 0.40$} & \small{$8.23\pm 0.13$}\\
\small{$\mathcal W_1$ ($\times 10^{-2}$) } & \small{$\mathbf{4.49\pm 0.54}$}
& \small{$116.1\pm 13.7 $}& \small{$23.58\pm 0.15 $}
& \small{$56.39\pm 5.73$} & \small{$10.36\pm 0.43$}\\
\small{$\mathcal W_2$ ($\times 10^{-2}$) } & \small{$\mathbf{0.39\pm 0.08}$}
& \small{$91.66\pm 15.36 $}& \small{$3.37\pm 0.06 $}
& \small{$22.25\pm 3.75$} & \small{$0.97\pm 0.09$} 
\end{tabular}
\end{center}
\end{table}

\begin{table}[th!]
\caption{All metrics M\"obius-to-circle: forward direction (global)}
\label{table:mob-to-circle-fwd}
\begin{center}
\begin{tabular}{r|rrrrr}
 &  \small{BMNet } & \small{AugCGAN } &\small{ LNFMM } & \small{cGAN }   & \small{cNF } 
 \\ \hline 
\small{MSMD ($\times 10^{-4}$) } & \small{$2.93\pm 1.82$}
& \small{$\mathbf{2.78\pm 0.36}$}& \small{$160.3\pm 97.2 $} &
\small{$136.8\pm 30.4$} &  \small{$49.13\pm 23.95$}\\
\small{MMD ($\times 10^{-3}$) }& \small{$0.37\pm 0.16$} 
& \small{$\mathbf{0.34\pm 0.23} $}& \small{$2.07\pm 1.13 $}
& \small{$4.55\pm 1.65$} &  \small{$1.42 \pm 0.74$}\\
\small{KL-fwd } & \small{$\mathbf{1.29\pm 0.51}$} 
& \small{$1.65\pm 0.11 $}& \small{$3.75\pm 0.17 $}
& \small{$4.56\pm 0.42$} & \small{$\mathbf{1.29\pm 0.03}$}\\
\small{KL-bwd } &\small{$3.83\pm 0.57$} 
& \small{$4.20\pm 0.09 $}& \small{$6.35\pm 0.21 $}
& \small{$7.58\pm 0.62$} &  \small{$\mathbf{3.78\pm 0.08}$}\\
\small{$\mathcal W_1$ ($\times 10^{-2}$) }& \small{$1.85\pm 0.39$} 
& \small{$\mathbf{0.68\pm 0.11} $}& \small{$6.50\pm 2.62 $}
& \small{$8.49\pm 1.54$} & \small{$2.58\pm 0.59$}\\
\small{$\mathcal W_2$ ($\times 10^{-2}$) }& \small{$0.14\pm 0.03$} 
& \small{$\mathbf{0.10\pm 0.06}$}& \small{$0.69\pm 0.46 $}
& \small{$1.308\pm 0.372$} & \small{$0.14\pm 0.14$} 
\end{tabular}
\end{center}
\end{table}

\begin{table}[th!]
\caption{All metrics M\"obius-to-circle: forward direction (local)}
\label{table:mob-to-circle-fwd}
\begin{center}
\begin{tabular}{r|rrrrr}
 &  \small{BMNet } & \small{AugCGAN } &\small{ LNFMM } & \small{cGAN }   & \small{cNF } 
 \\ \hline 
\small{MSMD ($\times 10^{-4}$) } &  \small{$\mathbf{5.68\pm 1.70}$} 
& \small{$6074\pm 2153 $}& \small{$204.2\pm 64.8 $}
& \small{$203.7\pm 60.6$} &  \small{$56.49\pm 22.10$}\\
\small{MMD ($\times 10^{-3}$) }&  \small{$14.69\pm 3.23$} 
& \small{$394.3\pm 67.8 $}& \small{$35.71\pm 5.55 $}
& \small{$184.1\pm 24.4$}&  \small{$\mathbf{13.37\pm 3.50}$}\\
\small{KL-fwd }  & \small{$1.83\pm 0.24$} 
& \small{$6.95\pm 1.02 $}& \small{$3.98\pm 0.31 $}
& \small{$4.58\pm 0.42$}&  \small{$\mathbf{0.66\pm 0.14}$}\\
\small{KL-bwd } & \small{$4.22\pm 0.31$} 
& \small{$10.64\pm 0.44 $}& \small{$6.21\pm 0.39 $}
& \small{$8.25\pm 0.35$}&  \small{$\mathbf{2.30\pm 0.21}$}\\
\small{$\mathcal W_1$ ($\times 10^{-2}$) } & \small{$9.48\pm 0.98$} 
& \small{$87.3\pm 13.41 $}& \small{$16.67\pm 2.08 $}
& \small{$42.63\pm 3.71$}&  \small{$\mathbf{6.08 \pm 0.93}$}\\
\small{$\mathcal W_2$ ($\times 10^{-2}$) }&  \small{$1.26\pm 0.27$} 
& \small{$62.3\pm 14.5 $}& \small{$3.29\pm 0.65 $}
& \small{$15.06\pm 2.26$} &  \small{$\mathbf{1.07\pm 0.28}$} 
\end{tabular}
\end{center}
\end{table}

\begin{table}[th!]
\caption{All metrics M\"obius-to-circle: reverse direction (global)}
\label{table:mob-to-circle-bwd}
\begin{center}
\begin{tabular}{r|rrrrr}
 &  \small{BMNet } & \small{AugCGAN } &\small{ LNFMM } & \small{cGAN }   & \small{cNF } 
 \\ \hline 
\small{MSMD ($\times 10^{-4}$) } & \small{$\mathbf{4.39\pm 1.61}$} &
 \small{$13.39\pm 2.12 $}& \small{$562.6\pm 543.5 $} &
\small{$6581\pm 561$} & 
 \small{$39.31\pm 1.91$}
 \\
\small{MMD ($\times 10^{-3}$) }& \small{$\mathbf{0.57\pm 0.11}$} 
& \small{$0.36\pm 0.13 $}& \small{$19.64\pm 16.99 $}
& \small{$126.3\pm 8.5$} & \small{$0.88\pm 0.23$}\\
\small{KL-fwd } & \small{$\mathbf{0.09\pm 0.05}$} 
& \small{$0.36\pm 0.03 $}& \small{$14.09\pm 1.47 $}
& \small{$12.95\pm 0.81$}  & \small{$0.28\pm 0.06$}\\
\small{KL-bwd } &\small{$\mathbf{0.94\pm 0.29}$} 
& \small{$1.29\pm 0.05 $}& \small{$7.87\pm 0.96 $}
& \small{$11.06\pm 0.23$} & \small{$1.04\pm 0.11$}\\
\small{$\mathcal W_1$ ($\times 10^{-2}$) }& \small{$\mathbf{2.33\pm 0.19}$} 
& \small{$2.37\pm 0.16 $}& \small{$23.89\pm 11.95 $}
& \small{$103.7\pm 4.1$} &  \small{$3.36\pm 0.05$}\\
\small{$\mathcal W_2$ ($\times 10^{-2}$) }& \small{$0.15\pm 0.03$} & 
 \small{$\mathbf{0.01\pm 0.01} $}& \small{$0.23\pm 1.16 $} &
\small{$61.24\pm 5.11$} &  \small{$0.13\pm .10$} 
\end{tabular}
\end{center}
\end{table}

\begin{table}[th!]
\caption{All metrics M\"obius-to-circle: reverse direction (local)}
\label{table:mob-to-circle-bwd}
\begin{center}
\begin{tabular}{r|rrrrr}
 &  \small{BMNet } & \small{AugCGAN } &\small{ LNFMM } & \small{cGAN }   & \small{cNF } 
 \\ \hline 
\small{MSMD ($\times 10^{-4}$) } & 
\small{$\mathbf{18.95\pm 1.32}$} 
& \small{$7230\pm 2280 $}& \small{$91.96\pm 44.44 $} 
& \small{$5652\pm 2396$} & 
\small{$92.58\pm 14.33$}
 \\
\small{MMD ($\times 10^{-3}$) }& \small{$12.81\pm 2.76$}
& \small{$386.5\pm 64.2 $}& \small{$126.7\pm 18.6 $}
& \small{$503.1\pm 24.1$} & 
\small{$\mathbf{8.12\pm 1.01}$}\\
\small{KL-fwd } & \small{$0.71\pm 0.11$} 
& \small{$8.02\pm 1.15 $}& \small{$26.44\pm 0.53 $} &
\small{$12.08\pm 1.06$}& 
\small{$\mathbf{0.35\pm 0.06}$}\\
\small{KL-bwd }  & \small{$0.98\pm 0.14$} 
& \small{$8.97\pm 0.59 $}& \small{$7.97\pm 0.13 $}
& \small{$8.67\pm 0.24$}& 
\small{$\mathbf{0.87\pm 0.08}$}\\
\small{$\mathcal W_1$ ($\times 10^{-2}$) }& \small{$8.77\pm 0.79$}
& \small{$94.32\pm 14.2 $}& \small{$42.78\pm 2.03 $}
& \small{$117.17\pm 1.84$}& 
\small{$\mathbf{8.60\pm 0.38}$}\\
\small{$\mathcal W_2$ ($\times 10^{-2}$) }& \small{$1.10\pm 0.24$}
& \small{$70.97\pm 16.03 $}& \small{$12.98\pm 1.34 $}
& \small{$74.42\pm 2.67$} & 
\small{$\mathbf{0.79\pm 0.09}$} 
\end{tabular}
\end{center}
\end{table}

\subsection{Review: Fiber Bundles}
\label{sect-fiber-bundles}

A \emph{fiber bundle} is a 4-tuple $(E,B,Z,\pi)$ of topological spaces and a (surjective) continuous map $\pi:E\to B$. $E$ is assumed to be covered by a collection of open sets $\{U_i\}_{i\in\mathcal I}$ and each $U_i$ has a homeomorphism $\varphi_i: \pi^{-1}(U_i) \rightarrow U_i \times Z$, called a {\emph{local trivialization}}, associated with it. Informally, $\varphi_i$ ensures that each neighborhood $\pi^{-1}(U_i)$ looks like a direct product of $U_i$ and $Z$. Finally, each $\varphi_i$ should respect the projection $\pi$ so that the following diagram commutes.

\begin{equation}
\label{eqn-comm-diagram}
    \begin{tikzpicture}
    \node at (0,0) {$\pi^{-1}(U_i)$};
    \node at (2.5,0) {$U_i \times Z$};
    \node at (0,-1.2) {$U_i$};
    
    \draw[->,black] (.8,0)--(1.8,0);
    \draw[->,black] (2.3,-.2)--(.3,-1);
    \draw[->,black] (0,-.2)--(0,-1);
    
    \node at (1.3,.28) {\scriptsize{$\varphi_i$}};
    \node at (1.3,-.9) {$\text{proj}$};
    \node at (-.2,-.6) {$\pi$};
    \end{tikzpicture}
\end{equation}

$E$ is known as the {\emph{total space}}, $B$ is known as the {\emph{base space}}, and $Z$ is known as the {\emph{fiber}}. Note that the product space $X \cong Y \times Z$ is trivially identified as a fiber bundle where $X$ is the total space $E$, $Y$ (respectively $Z$) is the base space $B$, $Z$ (resp. $Y$) is the fiber, $\pi$ is the usual projection map from $Y \times Z$ to $Y$ (resp. $Z$), $\{U_i\}_{i \in \mathcal{I}}$ consists of a single neighborhood which is all of $Y$ so that $\pi^{-1}(Y) = Y \times Z$, and $\varphi_i$ is just the identity map. The reader can check that diagram \eqref{eqn-comm-diagram} easily commutes in this case.

\end{document}